# Contribution of the Temperature of the Objects to the Problem of Thermal Imaging Focusing

Virginia Espinosa-Duró
Tecnocampus TCMM
Mataró, Barcelona, SPAIN
espinosa@eupmt.es

Marcos Faundez-Zanuy
Tecnocampus TCMM
Mataró, Barcelona, SPAIN
faundez@eupmt.es

Jiří Mekyska
DT, FEEC, BU
Brno, CZECH REPUBLIC
xmekys01@stud.feec.vutbr.cz

*Abstract*— When focusing an image, depth of field, aperture and distance from the camera to the object, must be taking into account, both, in visible and in infrared spectrum. Our experiments reveal that in addition, the focusing problem in thermal spectrum is also hardly dependent of the temperature of the object itself (and/or the scene).

## I. Introduction

As well as in visible image acquisition systems, an optical system capable of focusing all rays of light from a point in the object plane to the same point in the focal plane is desired, the same goal is also desired when dealing with thermal imagers, to get clear and focused images in thermal infrared spectrum. However, optical properties usually depend on wavelength, leading to major kind of lens *aberrations* as well as deviations due to the *diffraction* becoming more present in infrared light, drastically reducing the capability to focus IR images. In this section we will deal this problem beyond the visible spectrum.

We will firstly consider the *achromatic aberration*, as the most critical aberration aspect when dealing with broadband spectrum images which is the case of MWIR and LWIR thermal images. Secondly, and not least, we will focus on diffraction effects, highly dependent with the wavelength, that also spread the image, even if the optical system is free of any kind of aberration (chromatic, spherical, astigmatism…). The challenge becomes more difficult to solve when more than one object located at different distances in scene demands to be focused, especially due to the constraints in depth of field design [1].

### A. Chromatic aberration

Chromatic aberration is an undesirable optical effect that promotes the inability of the lens to focus all different wavelengths of the beam light (-all colors in visible light-) to the same focal point. This effect, sometimes also called achromatism or chromatic distortion, is due to the spread dispersion phenomenon concerning the refractive index variation with wavelength. Normal lens show normal dispersion, that is, the index of refraction n, decrease with increasing wavelength [2]. Thus, the light beam with longer wavelength is refracted less than the shorter wavelength one. This behavior produces a set of different focal points, as can be seen in Figure 1.

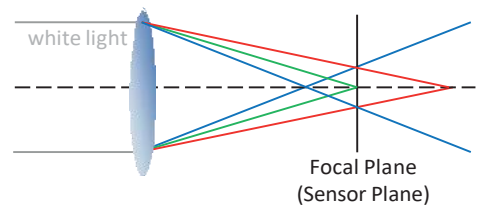

Fig. 1. Optical representation of chromatic aberration. Note that longer wavelengths focus long, while smaller ones, focuses before the focal plane.

Figure 2 shows a comparative of the same scene acquired with the same camera (sensitive to both, visible and Near-IR spectrum) and provided with a standard lens (not an infrared corrected one). Note that picture in NIR spectrum (Fig.2 on the right) is slightly out of focus due to the achromatic aberration.

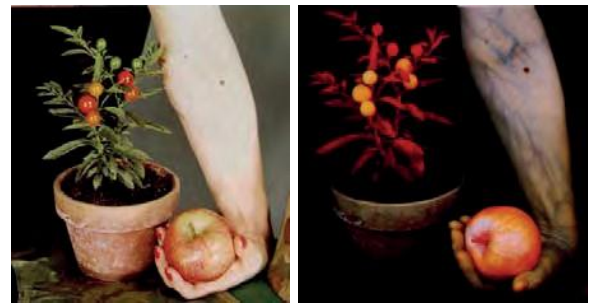

Fig. 2. Same scene taken in VIS (a) and NIR (b) spectrums. Focusing differences can be appreciated. In addition, strong differences in human skin reproduction can be viewed, as well as the fruit color reproduction. Extracted from [3].

In order to reduce the impact of such aberration, straightforward approaches in visible images exists, such as concentrating the light beam by setting small apertures (implies larger depths of field, DOF). Special low and extra low dispersion glasses which present a small variation of refractive index with wavelength are also available. Additionally, a number of high performance optical lenses also exist such as achromatic and apochromatic ones, for most demanding imaging corrections [2]. In any case, the correction tasks in visible spectrum are reasonable to achieve due to the short range of wavelengths to deal with. This is not the case in the MWIR and LWIR operating ranges, where thermal IR sensors measure simultaneously over broadband wavelength. Thus, while the change is 400nm between the violet and red end of the EM spectrum, in the both MWIR and LWIR spectra the wavelength ranges are 2000nm and 6000nm, respectively [4].

### B. Diffraction effect

Diffraction is an optical effect, due to the wave nature of light, which can limit the total resolution of any image acquisition process. Usually, light propagates in straight lines through air. However, this behavior is only valid when the wavelength of the light is much smaller than the size of the structure to through. For smaller structures, such a gap or a small hole, which is the case of camera's aperture, light beam will suffer a diffraction effect caused by a slight bending of light when it passes through such singular structures. Due to this effect, any image formed by a perfect optical lens of a point of light, not correspond to a point, but to a circle called Airy disc (not to be confused with the circle of confusion), and determines maximum blur allowable by the optical system. Furthermore, the diameter of this circle will be used to define the theoretical maximum spatial resolution of the sensor and will be given by the following equation:

$$2,44\lambda \frac{d}{D}$$

Where $\lambda$ is the light wavelength, d is the distance from the image to the lens and D is the effective aperture diameter. Solutions in this field only consider diffraction limited optical systems that provide a measure of the diffraction limit of a system. Knowing this limit can help you to avoid any subsequent softening.

### C. Problem Statement

In addition with both parameters referred, especially the chromatic aberration, the temperature of the object of the scene and the concerned heat transfer through the boundary between two systems (object and surrounding) [5], will be also taking into account in order to obtain clear, crisp thermal images.

## II. MATERIAL AND METHODS

In order to analyze the focus of a thermal image, we have constructed a special purpose database. Using this database, we have evaluated the focusing measure for each image and plotted it against the focus position.

### A. Acquisition System

Thermal images have been acquired using a thermographic camera TESTO 882, equipped with a silicon uncooled microbolometer detector with a spectral sensitivity range from 8 to 14µm and provided with a germanium optical lens. This thermal imager also integrates a 640x480 resolution AF visible camera. The key technical characteristics are summarized as follows:

- Sensor:
  - Type: UFPA, temperature stabilized.
  - Resolution: 160x120 pixels (320 x 240 pixels interpolated)
  - Spectral Sensitivity: 8 to 14µm
  - Thermal Sensitivity (NETD): <100mK at 30°C
- Removable angular optical lens:
  - FOV: 32°x23°; Focal Distance: 15mm; Fixed aperture: f/0,95
  - IFOV: 1,7mrad
  - Closest Focusing Distance: 10cm

### B. Database description

The database consists of eight image subsets of the same scene as showed in Figure 3 at eight different fixed temperatures of the bulb. Thus, each subset consists of 96 different images of the scene taken at an ambient temperature of 20 degrees

Bulb has been chosen in order to firstly approach the human face (It is a similar 3D object with the possibility to depict the same temperature than a human body). In this respect, bulb's temperature has been regulated by means of its current by using the proposed dimmer D1KS 220-240V 50-60Hz 1000W, obtaining a final span of temperatures near to 40°C (from 40°C to 80°C). In each set, the camera acquires one image at each lens position following the process defined in our previous work [4]. Thus, we have again manually moved the lens in 1mm steps which provides a total of 96 positions. For this purpose, we have attached a millimeter tape to the objective, as showed in Figure 2 and used a stable tripod in order to acquire the same scene for each scene position. The final performed database consists of 8x96 = 768 thermal images. Figure 3 show a sample of the best focused image of each subset.

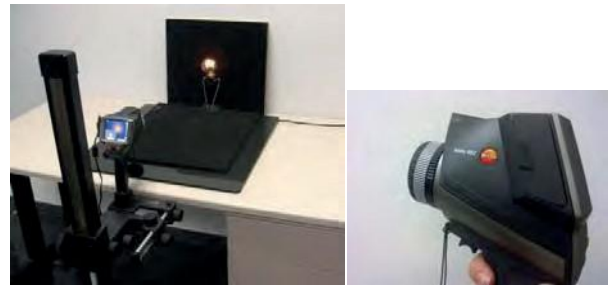

Fig. 3. (a) Scenario. (b) TESTO 882 thermal imager with the stepping ring manual adapter.

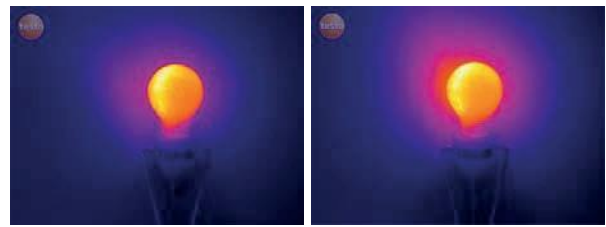

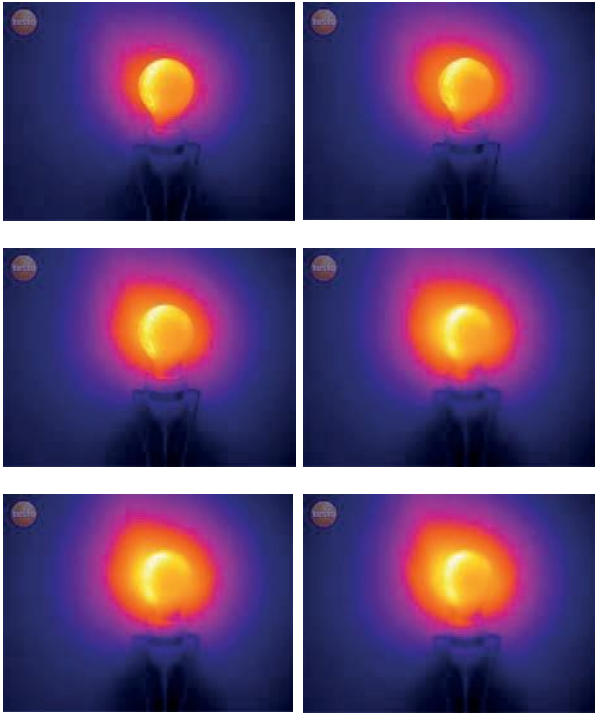

Fig. 4. Best image of each set, at the eight evaluated temperatures.

## C. Focus Measures.

In our previous work [4] we compared some approaches for image focus measuring. According to results presented, sum-modified Laplacian (SML) has been selected. Nayar and Nakagawa [6] noted that, in the case of the Laplacian, the second derivatives in the x- and y-directions can have opposite signs and tend to cancel each other out. Therefore, they proposed SML, which can be obtained by means of:

$$SML = \sum_{i=x-W}^{x+W} \sum_{j=y-W}^{y+W} \nabla^2_{ML} f(i,j) \quad for \nabla^2_{ML} f(i,j) \geq T$$

where *T* is a discrimination threshold value and:

$$A_{ML}^2 f(x,y) = |2I(x,y)-I(x-step,y)-I(x+step,y)| + |2I(x,y)-I(x,y-step)-I(x,y+step)|$$

In order to accommodate for possible variations in the size of texture elements, Nayar and Nagagawa used a variable spacing (step) between the pixels to compute ML. The parameter W determines the window size used to compute the focusing measure.

## III. EXPERIMENTAL RESULTS AND CONCLUSIONS

In this section we present the experimental results with the full database described in Section 2. By one side, is direct to obtain by means of the Wien's displacement law, that the bulb's temperature range of 40°C implies a second span of wavelengths of 1048nm that is more than two and a half times the fully visible spectrum, producing optical distortions due to the achromatic aberration. By other side, and taking into account that the temperature of the bulb's surrounding is about 20°C while the temperature and the concerned gradient is increasing in each subset, the conduction heat transfer phenomenon (also called diffusion), becomes more evident, as states the second law of thermodynamics. This involves solving the equations that describe the phenomenon characteristics, which result from the principles of conservation of mass, momentum and energy [5], being the referred analysis beyond the scope of this research.

For our study, determine and analytically verify, as sampled in Figure 5, that the larger the conduction heat transfer, the smaller the sharpness, (producing as in the analyzed case a glare around the hotter bulbs that dramatically contributes to the defocus of themselves), will be sufficient for the ultimate purpose of demonstrating that these both effects can be considerate negligible when dealing with facial thermograms (with temperatures near to 37°), being such features reliable and robust for biometric recognition purposes.

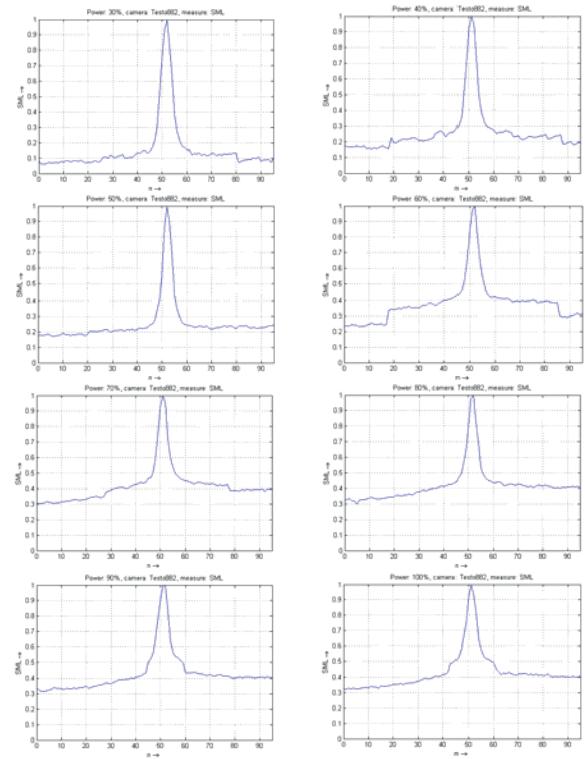

Fig. 5. Focusing measures of each of the 96 images per subset. The images that depict more heater bulbs depict less sharpener peaks. Said in another way, higher temperature, involves worse focusing results.

ACKNOWLEDGMENT

This work has been supported by FEDER, MEC and TEC2009-14123-C04-04. We also want to acknowledge the COST OC08057 project for providing Jiri's support.

REFERENCES

[1] Radek Benes, Pavel Dvorak, Marcos Faundez-Zanuy, Virginia Espinosa-Duró, Jiri Mekyska, "Multi-focus thermal image fusion", Pattern Recognition Letters, Volume 34, Issue 5, 2013, Pages 536-544, ISSN 0167-8655, https://doi.org/10.1016/j.patrec.2012.11.011

___________________________________________________________________________________________________________________

[2] R. Jacobson, S. Ray, G. Attridge and N. Axford, "Manual of Photography", Ninth Edition. Focal press. 2000.

[3] International Edition of the Encyclopedia of Practical Photography. Eastman Kodak Company Inc. American Photographic Book Publishing Company Inc. Editions Grammont, S.A. 1979. ISBN 84-345-3949-7.

[4] M. Faúndez-Zanuy, J. Mekiska and V. Espinosa-Duró, "On the Focusing of Thermal Images". Pattern Recognition Letters. Ed. Elsevier. Vol. 32. pp 1548-1557. August 2011. DOI: 10.1016/j.patrec.2011.04.022

[5] F.P. Incropera, "Fundamentals of Heat and Mass Transfer". 4[th] Edition. John Wiley & Sons, Inc, New York.

[6] K. Nayar and Y. Nakagawa, "Shape from Focus," IEEE Transactions on Pattern Analysis and Machine Intelligence, Vol. 16, Issue 8, pp. 824-831, August 1994.